\documentclass{article} 
\usepackage{nips15submit_e,times}
\usepackage{hyperref,amsmath,graphicx,amsfonts}
\usepackage{url}

\title{Information Theoretic-Learning Auto-Encoder}

\author{
Eder Santana
\thanks{The companion source code of this paper
can be found online: https://github.com/cnel/itl-ae} \\
University of Florida\\
\And
Matthew Emigh \\
University of Florida\\
\AND
Jose C. Principe \\
University of Florida\\
}

\nipsfinalcopy 

\begin{document}

\maketitle

\begin{abstract}
We propose Information Theoretic-Learning (ITL) divergence measures for
    variational regularization of neural networks. We also explore
    ITL-regularized autoencoders as an alternative to variational
    autoencoding bayes, adversarial autoencoders and generative adversarial
    networks for randomly generating sample data without explicitly defining a
    partition function. This paper also formalizes, generative moment matching
    networks under the ITL framework.
\end{abstract}

\section{Introduction}
\label{sec:intro}

Deep, regularized neural networks work better in practice than shallow
unconstrained neural networks \cite{deepbook}. This regularization takes classic forms
such as L2-norm ridge regression, L1-norm LASSO, architectural
constraints such as convolutional layers \cite{lecun1998gradient}, but also uses modern
techniques such as dropout \cite{srivastava2014dropout}. Recently,
especially in the subfield of autoencoding
neural networks, regularization has been accomplished with variational methods 
\cite{VAE}\cite{rezende2014stochastic}.
In this paper we propose Information Theoretic-Learning \cite{itlbook} divergence measures
for variational regularization.

In deep learning, variational regularization forces the function implemented by
a neural network to be as close as possible to an imposed prior,
which is a stronger restriction than that imposed by point-wise
regularization methods such as L1 or L2 norms.
Variational methods for deep learning were popularized by the variational
autoencoder (VAE) framework proposed by \cite{VAE} and
\cite{rezende2014stochastic} which also brought
the attention of deep learning researchers to the reparametrization trick. The
Gaussian reparametrization trick works as follows: the encoder (deep)
network outputs a mean $\mu$ and a standard deviation $\sigma$, from which we
sample a latent factor $z=\mu + \sigma \cdot \epsilon$, where 
$\epsilon \sim N(0, 1)$. This latent factor is then fed forward to the
decoder network and
the parameters $\mu$ and $\sigma$ are regularized using the KL-divergence
$KL(N(\mu, \sigma) \| N(0, 1))$ between the inferred distribution and the imposed prior,
which has a simple form \cite{VAE}. After training,
one can generate data from a VAE by first sampling from
the Gaussian prior distribution and feeding it to the VAE's decoder. This is an approach
similar to the inverse cumulative distribution method and does not involve
estimation of the partition function, rejection sampling, or other complicated
approaches \cite{mcmcbook}.
VAE's methodology has been successfully extended to convolutional autoencoders
\cite{kulkarni2015deep} and more elaborate 
architectures such as Laplacian pyramids for image generation
\cite{denton2015deep}.

Unfortunately, VAE cannot be used when there does not exist a simple closed form solution for the KL-divergence. 
To cope with that, generative adversarial networks (GAN) were
proposed \cite{goodfellow2014generative}. GAN uses two neural networks that are trained competitively---a
generator network $G$ for sampling data and a discriminator network $D$
for discerning the outputs of $G$ from real data. Unfortunately, training $G$ to match a
high dimensional dataset distribution using only $D$'s binary ``fake'' or ``legit''
outputs is not a stable or simple process.  

Makhzani et. al. proposed
adversarial autoencoders \cite{makhzani2015adversarial} which use 
an adversarial discriminator $D$ to tell the low dimensional codes in the output of
the encoder from data sampled from a desired distribution. In this way adversarial
autoencoders can approximate
variational regularization as long as it is possible to sample from the desired
distribution. We note that although this partially
solves the problem of generalized functional regularization for neural
networks
\footnote{We still have a problem when we cannot sample from the desired distribution.},
adversarial autoencoders require us to
train a third network, the discriminator, in addition to the encoder and decoder already
being trained. 

Here we observe that, assuming we can sample from the desired distribution,
we can use empirical distribution divergence measures proposed by Information Theoretic-Learning
(ITL) as a measure of how close the function implemented by an encoder network is
to a desired prior distribution. Thus, we propose Information Theoretic-Learning Autoencoders (ITL-AE).

In the next section of this paper we review ITL's Euclidean and
Cauchy-Schwarz divergence measures \cite{principe2000information}.
In Section 3 we propose the ITL-AE and 
run experiments to illustrate the proposed method in Section 4. We conclude the paper
afterwards.

\section{Information Theoretic-Learning}
\label{sec:itl}
Information-theoretic learning (ITL) is a field at the intersection of machine
learning and information theory \cite{itlbook} which encompasses a family of algorithms that
compute and optimize information-theoretic descriptors such as entropy,
divergence, and mutual information. ITL objectives are computed directly from samples
(non-parametrically) using Parzen windowing and Renyi's entropy
\cite{renyi1961measures}.

\subsection{Parzen density estimation}
Parzen density estimation is a nonparametric method for estimating the pdf of a
distribution empirically from data. For samples $x_i$ drawn from a distribution
$p$, the parzen window estimate of $p$ can be computed nonparametrically as 
\begin{equation}\label{eq:parzen}
\hat{p}(x)=\frac{1}{N}\sum_{i=1}^N G_\sigma (x - x_i).
\end{equation}

\begin{figure}[t]
\centering
\includegraphics[scale=0.37]{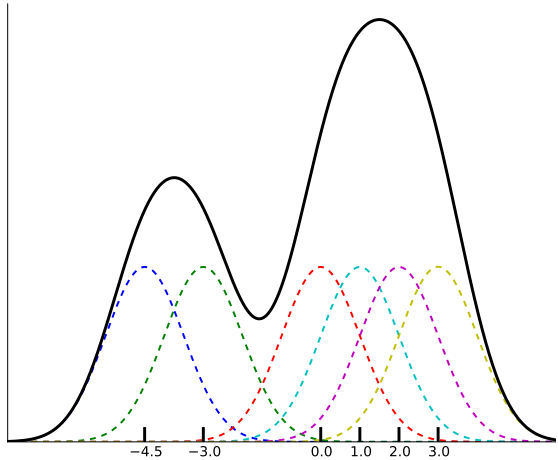}
\caption{One-dimensional Parzen windowing example with Gaussian kernel functions. A small Gaussian (colored dashed lines) is placed at each sample, the sum of which (solid black line) is used to estimate the pdf from which samples (vertical tick marks) were drawn.}
\label{fig:parzen}
\end{figure}
Intuitively, as shown in Fig \ref{fig:parzen}, parzen estimation corresponds to
centering a Gaussian kernel at each sample $x_i$ drawn from $p$, and then
summing to estimate the pdf. The optimal kernel size \cite{silverman1986density}
depends on the density of samples, which approaches zero as the number of
samples approaches infinity.

\subsection{ITL descriptors}
 Renyi's $\alpha$-order entropy for probability density function (pdf) $p$ is given by:
\begin{equation} \label{eq:renyi_entropy}
H_\alpha (X) = \frac{1}{1-\alpha} \log \int p^\alpha(x) dx
\end{equation}
where $p \in L^\alpha$. Renyi's $\alpha$-order entropy can be considered a
generalization of Shannon entropy  since $\lim_{\alpha \to 1}H_\alpha = \int
-\log p(x) dx$, which is Shannon entropy. For the case of $\alpha=2$, equation
\eqref{eq:renyi_entropy} simplifies to $H_2 = -\log \int p^2(x) dx$ which is
known as Renyi's quadratic entropy.

Plugging \eqref{eq:parzen} into \eqref{eq:renyi_entropy}, for $\alpha=2$, we obtain: 
\begin{align}
\hat{H}_2(X) &= -\log \int \left( \frac{1}{N} \sum_{i=1}^N G_\sigma (x - x_i) \right)^2 dx \\
&= -\log \left( \frac{1}{N^2} \sum_{i=1}^N \sum_{j=1}^N G_{\sigma \sqrt{2}} (x_j - x_i) \right) \label{renyi_estimator}
\end{align}
where $G_\sigma(x,y) = \frac{1}{\sqrt{2\pi} \sigma} \exp \left(
\frac{\|x-y\|^2}{2\sigma^2} \right)$ is the Gaussian kernel, and $\sigma$ is the
kernel size. The argument of the logarithm in equation \eqref{renyi_estimator}
is called the \emph{information potential} by analogy with potential fields from
physics and is denoted by $\hat{V}_\sigma(X)$.

Another important ITL descriptor is Renyi's cross-entropy which is given by:
\begin{equation} \label{eq:cross_ent}
H_2(X,Y) = -\log \int p_X(z)p_Y(z)dz 
\end{equation}
Similarly to equation \eqref{eq:renyi_entropy}, cross-entropy can be estimated by
\begin{equation} \label{eq:cross_est}
\hat{H}_2(X,Y) = -\log \frac{1}{N_X N_Y}\sum_{i=1}^{N_X} \sum_{j=1}^{N_Y} G_{\sqrt{2}\sigma} (x_i - y_j) 
\end{equation}
The argument of the logarithm in equation \eqref{eq:cross_ent} is called
\emph{cross-information potential} and is denoted $\hat{V}_\sigma(X,Y)$.
Cross-information potential can be viewed as the average sum of interactions of
samples drawn from $p_X$ with the estimated pdf $\hat{p}_Y$ (or vice-versa).  

ITL has also described a number of divergences connected with Renyi's entropy.
In particular, the Euclidean and Cauchy-Schwarz divergences are given by:
\begin{align} \label{eq:D_ED}
D&_{ED}(p_X\|p_Y) = \int \left( p_X(z) - p_Y(z) \right)^2 dz \\
& = \int p_X^2(z)dz + \int p_Y^2(z)dz - 2 \int p_X(z)p_Y(z)dz
\end{align}
and
\begin{equation} \label{eq:D_CS}
D_{CS}(p_X\|p_Y) = -\log \frac{\left( \int p_X(z) p_Y(z) dz \right)^2}{\int p_X^2(z) dz \int p_Y^2(z) dz},
\end{equation}
respectively.
Equations \eqref{eq:D_ED} and \eqref{eq:D_CS} can be put in terms of information potentials:
\begin{align} \label{eq:D_pot}
D_{ED}(p_X||p_Y) &= V(X) + V(Y) - 2 V(X,Y) \\
D_{CS}(p_X||p_Y) &= \log \frac{V(X) V(Y)}{V^2(X,Y)}
\end{align} 
Euclidean divergence is so named because it is equivalent to the Euclidean
distance between pdfs. Furthermore, it is equivalent to maximum mean discrepancy
\cite{gretton2006kernel} (MMD) statistical test. Cauchy-Schwarz divergence is named for the
Cauchy-Schwarz inequality, which guarantees the divergence is only equal to zero
when the pdfs are equal almost everywhere. $D_{CS}$ is symmetric but, unlike
$D_{ED}$, does not obey the triangle equality.

Minimizing either divergence over $p_X$, i.e., $\min_{p_X} D(p_X \|p_Y)$, 
is a tradeoff between minimizing the information potential (maximizing
entropy) of $p_X$ and maximizing the cross-information potential (minimizing
cross-entropy) of $p_X$ with respect to $p_Y$. Intuitively, minimizing the
information potential encourages samples from $p_X$ to spread out, while
maximizing the cross-information potential encourages samples from $p_X$ to move
toward samples from $p_Y$.

\section{ITL Autoencoders}
\label{sec:itlae}
Let us define autoencoders as a 4-tuple $AE = \{E, D, L, R\}$. Where $E$ and
$D$ are the encoder and the decoder functions, here parameterized as neural
networks. $L$ is the reconstruction cost function that measures the difference
between original data samples $x$ and their respective reconstructions
$\tilde{x} = D(E(x))$. A typical reconstruction cost is mean-squared error.
$R$ is a functional regularization. Here this functional
regularization will only be applied to the encoder $E$. Nonetheless, although 
we are only regularizing the encoder $E$, the
interested investigator could also regularize another intermediate layer of the
autoencoder \footnote{For those interested in such investigation we recommend
modifying the companion code of this paper. In our method adding more regularization does not
increase the number of adaptive weights.}.

The general cost function for the ITL-AE can be summarized by the following equation:
\begin{equation}
\text{cost} = L \left(x,\tilde{x} \right) + \lambda R(E,P),
\end{equation}
where the strength of regularization is controlled by the scale parameter $\lambda$, and $P$ is the imposed prior.

\begin{figure}[!ht]
\centering
\includegraphics[width=0.5\textwidth]{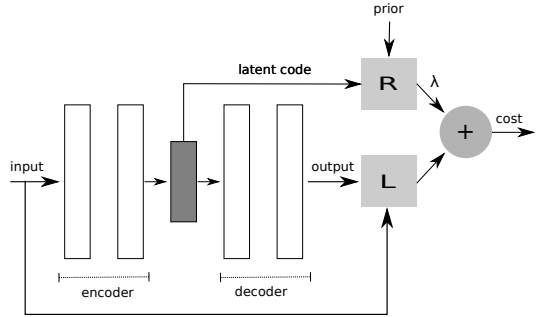}
\caption{Block diagram for ITL autoencoder. L is the reconstruction cost
function while R is the functional regularization that uses information
theoretic measures. }
\label{fig:varreg}
\end{figure}

The functional regularization costs investigated in this paper are the ITL Euclidean
and Cauchy-Schwarz divergences.  Both types of divergence encourage a smooth
manifold similar to the imposed prior. That is, maximizing latent code distribution
entropy encourages code samples to spread out, while minimizing the
cross-entropy between the latent code and the prior distributions encourages
code samples to be similar to prior samples. 

Note that if the data dimensionality is too high, ITL divergence measures
require larger batch sizes to be reliably estimated, this explains why Li et.
al. \cite{li2015generative} used batches of 1000 samples in their experiments and also why they
reduced the data dimensionality with autoencoders. In our own experiments (not
shown), the Cauchy-Schwarz divergence worked better than Euclidean for
high-dimensional data.

\section{Relation to other work}
Generative Moment Matching Networks (GMMNs) \cite{li2015generative} correspond to the specific case
where the input of the decoder $D$ comes from a multidimensional uniform
distribution and the reconstruction function $L$ is given by the Euclidean
divergence measure. GMMNs could be applied to generate samples from the original
input space itself or from a lower dimensional previously trained stacked
autoencoder (SCA) \cite{bengio2012better} hidden space. An advantage of our
approach compared to GMMNs is that we
can train all the elements in the 4-tuple $AE$ together without the elaborate
process of training layerwise stacked autoencoders for dimensionality reduction.

Variational Autoencoders (VAE) \cite{VAE} adapt a lower bound of the variational regularization,
$R$, using parametric, closed form solutions for the KL-divergence. That divergence can be
defined using Shannon's entropy or $H_{\alpha=1}$. Thus, we can also interpret ITL-AE as
nonparametric variational autoencoders, where the 
likelihood of the latent distribution is estimated empirically using Parzen
windows. Note that since we can estimate that distribution directly, we
do not use the reparametrization trick. Here the reparametrization trick could
possibly be used for imposing extra regularization, just like how adding dropout
noise regularizes neural networks.

Adversarial autoencoders (AA) \cite{makhzani2015adversarial} have the
architecture that inspired our method the
most. Instead of using the adversarial trick to impose regularization on the
encoder, we defined that regularization from first principles, which
allowed us to train a competing method with much fewer trainable parameters.
Our most recent experiments show that AA scales
better than ITL-AE for high dimensional latent codes. We leave investigation
into high dimensional ITL-AE for future work.

\begin{figure*}[!ht]
\centering
\includegraphics[width=.7\textwidth]{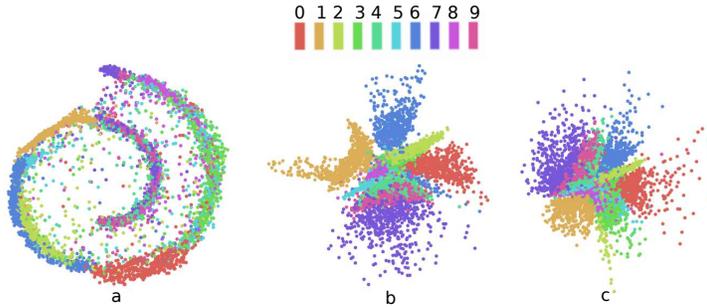}
\caption{Effect of different priors on the function defined by the encoder
neural network. a) Swiss-roll, b) Laplacian, b) Gaussian.}
\label{fig:varreg}
\end{figure*}

\section{Experiments}
\label{sec:exp}

\begin{figure}[!ht]
\centering
\includegraphics[width=0.3\textwidth]{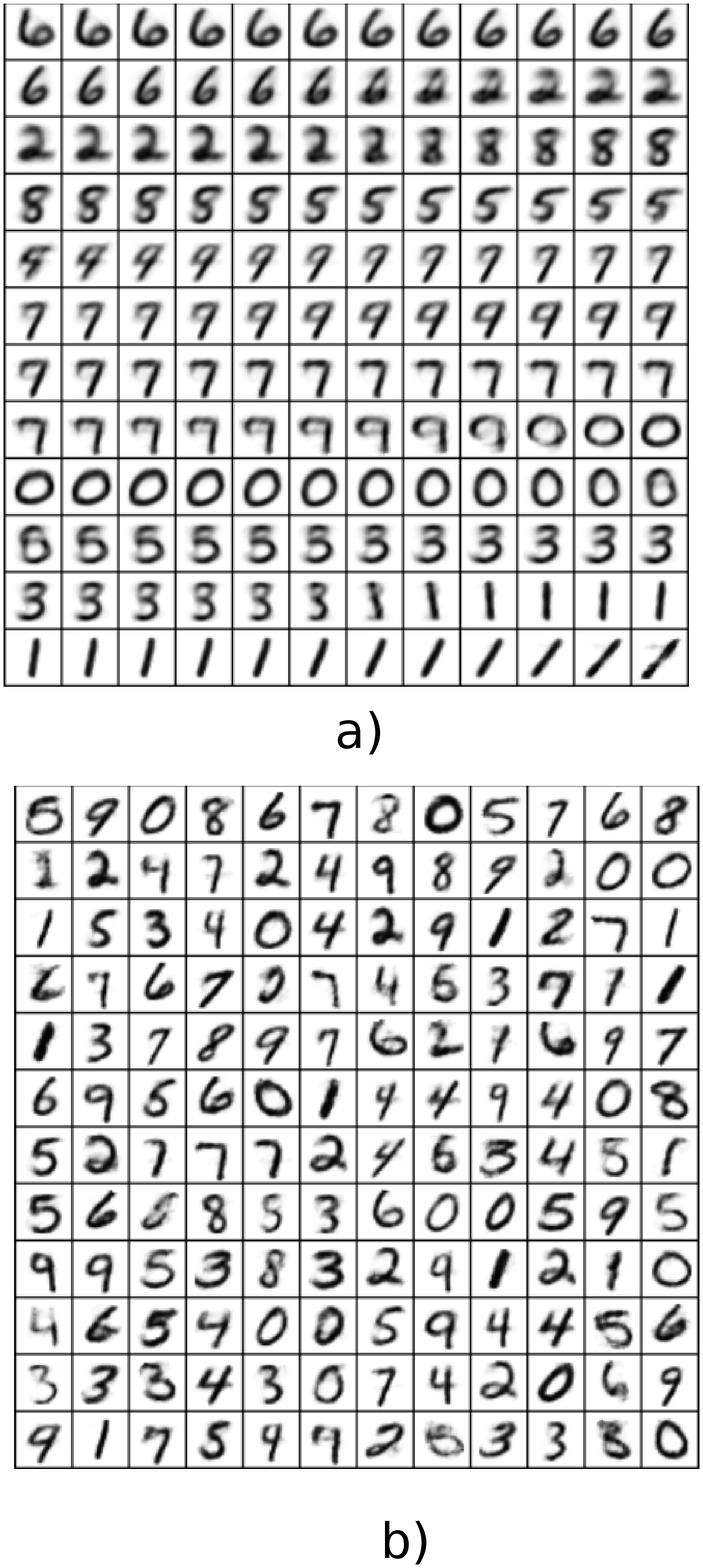}
\caption{Samples from the embedded manifolds. a) samples from a linear walk over
the swiss-roll manifold, b) random samples from a zero mean, std 5 Gaussian
distribution.}
\label{fig:samples}
\end{figure}

In this section we show experiments using the ITL-AE architecture described in the previous
section.  First, for a visual interpretation of the effects of variational
regularization we trained autoencoders with 2-dimensional latent codes. We used
as desired priors a Gaussian, a Laplacian, and a 2D swiss-roll distribution. The
resulting codes are shown in Fig. \ref{fig:varreg}. Note that in all of these
examples the autoencoder was trained in a completely unsupervised manner.
However, given the simplicity of the data and the imposed reconstruction cost,
some of the numbers were clustered in separate regions of the latent space.
Fig. \ref{fig:samples} shows some images obtained by sampling from a linear path
on the swiss-roll and random samples from the Gaussian manifold.

For easier comparisons and to avoid extensive hyperparameter search, we constrained
our encoder and decoders, $E$ and $D$, to have the same architecture as those
used in \cite{makhzani2015adversarial} i.e., each network is a two hidden layer
fully connected network with
1000 hidden neurons. Thus, the only hyperparameters investigated in this paper
were kernel size $\sigma$ and scale parameter $\lambda$. For the MNIST dataset,
the Euclidean distance worked better with smaller kernels, such as $\sigma=1$ or
$\sigma=5$, while the Cauchy-Schwarz divergence required larger kernel,
$\sigma=10$ for example. Nevertheless, here we will focus in regularizing the
low dimensional latent codes and leave experiments using Cauchy-Schwarz
divergence for future work.

Our best results for small batch sizes common in deep learning had 3-dimensional
latent codes in the output of the encoder, euclidean divergence as the
regularization $R$ and mean-squared error as the reconstruction cost $L$.  As we
will show in the next section, we were able to obtain competitive results and
reproduce behaviors obtained by methods trained with larger networks or extra
adversarial networks.

\subsection{Log-likelihood analysis}

We followed the log-likelihood analysis on the MNIST dataset reported
on the literature
\cite{hinton2006fast}\cite{bengio2012better}\cite{bengio2013deep}. After
training ITL-AE on the training set, we
generated $10^4$ images by inputting $10^4$ samples from a $N(0, 5)$ distribution to
the decoder.  Those generated MNIST images were used estimate a distribution
using Parzen windows on the high dimensional image space\footnote{Note that this
is not an optimal benchmark due the problems with Parzen estimators in high
dimensional spaces we explained. All the results, including ours, should be
taken with a grain of salt.}. We calculated the log-likelihood of separate 10k
samples from the test set and report the results on Table \ref{table:ll}. The
kernel size of that Parzen estimator was chose using the best results on a
held-out cross-validation dataset. That kernel size was $\sigma=0.16$. Note that
our method obtained the second best results between all the compared fully
connected generative models. Remember that this was obtained with about $10^6$
less adaptive parameters than the best method Adversarial autoencoders.   

\begin{table}[ht]
    \caption{Log-likelihood of MNIST test dataset. Higher the values are
    better.}
    \centering
    \begin{tabular}{c c c c}
    \hline
    Methods & Log-likelihood \\ [0.5ex] 
    \hline
    Stacked CAE \cite{bengio2012better}& $121 \pm 1.6$ \\
    DBN \cite{hinton2006fast}& $138\pm 2$ \\
    Deep GSN \cite{bengio2013deep}& $214\pm 1.1$ \\
    GAN \cite{goodfellow2014generative}& $225\pm 2$ \\
    GMMN + AE \cite{li2015generative}& $282\pm 2$ \\
    ITL-AE$^*$ & $300 \pm 0.5$  \\
    Adversarial Autoencoder \cite{makhzani2015adversarial}& $340 \pm 2$ \\ [1ex]
    \hline
    $^*$ Proposed method.
    \end{tabular}
    \label{table:ll}
\end{table}

\section{Conclusions}
\label{sec:conclusions}

Here we derived and validated the Information Theoretic-Learning Autoencoders, a
non-parametric and (optionally) deterministic alternative to Variational
Autoencoders. We also revisited ITL for neural networks, but this time, instead
of focusing on nonparametric cost functions for non-Gaussian signal processing,
we focused on distribution divergences for regularizing deep architectures.

Our results using relatively small, for deep learning standards, 4 layer networks
with 3-dimensional latent codes obtained competitive results on the log-likelihood
analysis of the MNIST dataset. 

Although our results were competitive for fully connected architectures, future
work should address the scalability of the ITL estimators for large dimensional
latent spaces, which is common on large neural networks and convolutional
architectures as well.

\subsubsection*{References}
\bibliographystyle{IEEEbib}

\end{document}